\documentclass[letterpaper]{article} 
\usepackage{aaai23}  
\usepackage{times}  
\usepackage{helvet}  
\usepackage{courier}  
\usepackage[hyphens]{url}  
\usepackage{graphicx} 
\usepackage{natbib}  
\usepackage{caption} 
\usepackage{algorithm}
\usepackage{algorithmic}

\usepackage{newfloat}
\usepackage{listings}
\DeclareCaptionStyle{ruled}{labelfont=normalfont,labelsep=colon,strut=off} 
\floatstyle{ruled}
\newfloat{listing}{tb}{lst}{}
\floatname{listing}{Listing}
\newcommand{\eg}{\emph{e.g., }} 
\usepackage{amsmath}  
\usepackage{amssymb}  
\usepackage{siunitx} 
\usepackage{multirow}
\usepackage{makecell}
\usepackage{subfigure}
\usepackage{color}

\title{Adaptive Dynamic Filtering Network for Image Denoising}

\author{
	Hao Shen\textsuperscript{\rm 1,2,4},
	Zhong-Qiu Zhao\textsuperscript{\rm 1,2,3,4}\thanks{Corresponding author.},
	Wandi Zhang\textsuperscript{\rm 1,2,4}
}
\affiliations{
	\textsuperscript{\rm 1}School of Computer Science and Information Engineering, Hefei University of Technology (HFUT) \\
	\textsuperscript{\rm 2}Intelligent Interconnected Systems Laboratory of Anhui Province (HFUT)\\
	\textsuperscript{\rm 3}Guangxi Academy of Sciences \\
	\textsuperscript{\rm 4}Intelligent Manufacturing Institute of HFUT\\
	{haoshenhs@gmail.com,z.zhao@hfut.edu.cn,wandizhang@mail.hfut.edu.cn}
}

\usepackage{bibentry}

\begin{document}
	\makeatletter
	\renewcommand{\@thesubfigure}{\hskip\subfiglabelskip} 
	
	\maketitle
	\begin{abstract}
		In image denoising networks, feature scaling is widely used to enlarge the receptive field size and reduce computational costs. This practice, however, also leads to the loss of high-frequency information and fails to consider within-scale characteristics. Recently, dynamic convolution has exhibited powerful capabilities in processing high-frequency information (\eg edges, corners, textures), but previous works lack sufficient spatial contextual information in filter generation. To alleviate these issues, we propose to employ dynamic convolution to improve the learning of high-frequency and multi-scale features. Specifically, we design a spatially enhanced kernel generation (SEKG) module to improve dynamic convolution, enabling the learning of spatial context information with a very low computational complexity. Based on the SEKG module, we propose a dynamic convolution block (DCB) and a multi-scale dynamic convolution block (MDCB). The former enhances the high-frequency information via dynamic convolution and preserves low-frequency information via skip connections. The latter utilizes shared adaptive dynamic kernels and the idea of dilated convolution to achieve efficient multi-scale feature extraction. The proposed multi-dimension feature integration (MFI) mechanism further fuses the multi-scale features, providing precise and contextually enriched feature representations. Finally, we build an efficient denoising network with the proposed DCB and MDCB, named ADFNet. It achieves better performance with low computational complexity on real-world and synthetic Gaussian noisy datasets. The source code is available at \url{https://github.com/it-hao/ADFNet}.	
	\end{abstract}
	\section{Introduction}
	Image denoising aims to recover the clean image from the observed noisy image, which is an essential step in improving image perception. In the early, most methods~\cite{K_SVD,zoran2011learning} were based on image priors derived from the statistics of natural images to remove noise. However, the performance degradation will appear once the noise distribution is inconsistent with the image priors. For the past few years, convolutional neural networks (CNNs), learning the mapping from the noisy image to the corresponding clean image in a data-driven manner, have been used to perform image denoising and have achieved superior performance. 
	
	\begin{figure}[t]
		\centering
		\includegraphics[width=0.47\textwidth]{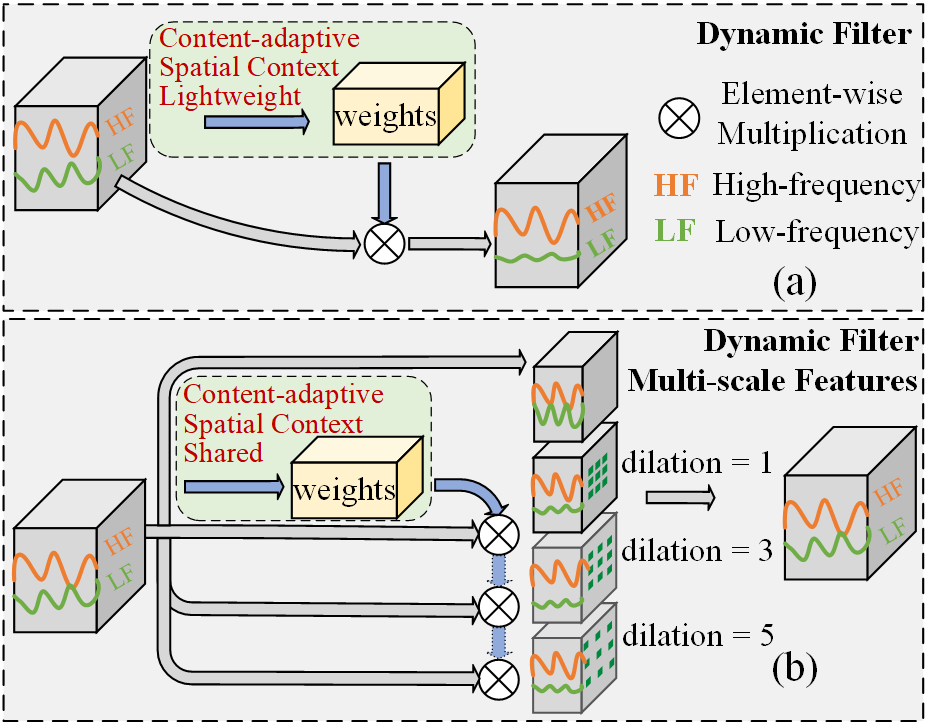} 
		\caption{Dynamic convolution and multi-scale dynamic convolution schemes.}
		\label{fig:motivation}
		\vspace{-0.4cm}
	\end{figure}
	
	Deep CNN-based methods~\cite{rnan,rdn} were recently proposed to expand the receptive field and pursue better performance. However, these methods also result in high computation consumption and time consuming, and could produce over-smoothed results with some high-frequency information (\eg edge, texture) lost. There are two reasons to explain it. First, regular static convolutions mainly focus on processing low-frequency information but are short of the capability to process high-frequency information~\cite{bi_volution}. Second, feature downsampling operation is widely used in deep CNNs to reduce parameters and computation. This practice, however, inevitably degenerates high-frequency signals, thus losing some informative image details~\cite{zou2020delving,agsn}. Compared to static convolution, dynamic convolution has distinct properties with spatial-anisotropy and content-adaptiveness, endowing the network with powerful capabilities to restore complex and delicate high-frequency information. The kernel generation module is an essential component of dynamic convolution, which predicts the corresponding kernels based on the input features. Therefore, this module should be highly efficient and lightweight to ensure dynamic convolution as a fundamental plug-and-play cell. As pioneer works, the kernel computation in \cite{hypernetworks, dfn} introduces a large number of parameters and is difficultly extended at deep CNN-based networks. \cite{involution} utilizes the depth-wise separable convolution manner to obtain relatively superiority in computational complexity but fails to use channel-specific information in kernel prediction. \cite{ddfnet} further design a channel filter branch to add channel-wise information to kernel generation. However, the absence of spatial information still remains. Furthermore, these works mainly focus on downstream tasks including detection and segmentation. Image denoising task requires removing noise and keeping fine-grained feature representation. Therefore, how to apply dynamic convolution to denoising networks and achieve fast and efficient denoising targets is a crucial issue.
	%
	
	Besides, multi-scale features have played an important role in the image denoising task, including two mainstream designs: global encoder-decoder architectures~\cite{sadnet} and local multi-scale feature extraction module~\cite{mprnet}. Both designs can enlarge the receptive field size of networks, thus being more semantics and robustness to noise. The former can extract cross-scale features and gradually recover high-resolution representation from coarse to fine. The latter can capture resolution-specific (within-scale) multi-scale feature representations with the help of various size contexts. However, almost existing studies cannot effectively incorporate these two manners, let alone inherit the merits of dynamic convolution and apply it to extract enriched multi-scale features. 
	
	In this work, we propose a novel Adaptive Dynamic Filtering Network (ADFNet) for image denoising, which incorporates dynamic convolution operation to tackle the above issues. To be specific, we propose a spatially enhanced kernel generation (SEKG) module to improve the dynamic convolution, realizing the interaction of spatial context information with slight additional computational complexity. The SEKG consists of spatial context extractor and channel information interaction branches, which can guide the generation of dynamic kernels, improving the high-frequency information reconstruction. Based on this module, we propose a dynamic convolution block (DCB) and a multi-scale dynamic convolution block (MDCB). The former enhances the high-frequency information via dynamic convolution (Fig.~\ref{fig:motivation} (a)) and preserves low-frequency information via skip connections. The latter expands the receptive field size by introducing the idea of dilated convolution and extracts powerful multi-scale features by applying shared adaptive dynamic kernels (Fig.~\ref{fig:motivation} (b)). The shared mechanism aims further to decrease the calculation of the whole network. In addition, a multi-dimension feature integration (MFI) mechanism is proposed to capture multi-scale feature interaction via cross-dimension inter-dependencies, endowing the network with a more powerful representative capability. When performing the dynamic convolution, we introduce the idea of depth-wise separable convolution to improve efficiency and reduce the model complexity. 
	Our main contributions are summarized as follows:
	
	\begin{itemize}
		\item We propose a spatially enhanced kernel generation (SEKG) module to implement dynamic convolution (DConv), which incorporates the learning of spatial context to kernel generation.
		\item With the guidance of the SEKG module, we propose a dynamic convolution block (DCB) and a multi-scale dynamic convolution block (MDCB) to enhance high-frequency and multi-scale feature representations. Extended experiments deploying DConv and MDCB to different backbone architectures show the effectiveness of the proposed SEKG scheme.
		\item We propose a new encoder-decoder network, named adaptive dynamic filtering network, using the proposed modules. Extensive experiments are conducted on real-world and synthetic Gaussian noisy datasets to show effectiveness and efficiency.
	\end{itemize}
	
	%
	%
	%
	%
	%
	
	\section{Related Work}
	\label{related_work}
	\textbf{CNN-based Image Denoising.} 
	Recently, many CNN-based methods have been proposed~\cite{jsnet,idr} and achieved state-of-the-art performance. For instance, \cite{dncnn} proposed to apply the residual learning and batch normalization to facilitate the training for image denoising. \cite{ffdnet} utilized the tunable noise level map to guide the recovery of the noisy images. Many following works further improve the performance by deploying elaborate architecture designs, including dense connections~\cite{rdn}, encoder-decoder architecture~\cite{red}, non-local attention~\cite{rnan}, dilated convolution~\cite{sadnet}, multi-scale design~\cite{msanet}, and others~\cite{deamnet}. However, many of these approaches adopt static convolutions to extract features, which can cause over-smoothing artifacts. And most of them have huge network structures, leading to a large amount of computation and low inference speed. Different from the above methods, our ADFNet adopts dynamic convolution to implement high-frequency and multi-scale feature extraction and achieves competitive results with acceptable model complexity.
	\newline
	\textbf{Dynamic Filtering.} 
	In contrast to static convolution with fixed kernels, dynamic convolution considers the image content and spatial position and can be categorized into two types. One kind~\cite{context,dconv,omni} is to predict the coefficients of different convolution kernels to combine them dynamically. The other is to utilize a separate network branch to predict kernels applied to the target features. The former only increases additional computational complexity to original static convolution kernels and is therefore difficult to apply to deep networks. Regarding the latter category, \cite{involution} utilized the efficient depth-wise separable convolution manner to implement the dynamic convolution operator but failed to encode channel-specific information. \cite{ddfnet} developed a channel filter branch, adding channel-wise information via squeeze-and-excitation structure to dynamic convolution kernels in an efficient way. However, the spatial context information cannot be fused effectively into kernel generation. Therefore, the proposed spatially enhanced kernel generation module fully collects the spatial and channel context information to guide the kernel generation.  
	\section{Method}
	\subsection{Network Structure}
\begin{figure}[t]
		\centering
		\includegraphics[width=0.45\textwidth]{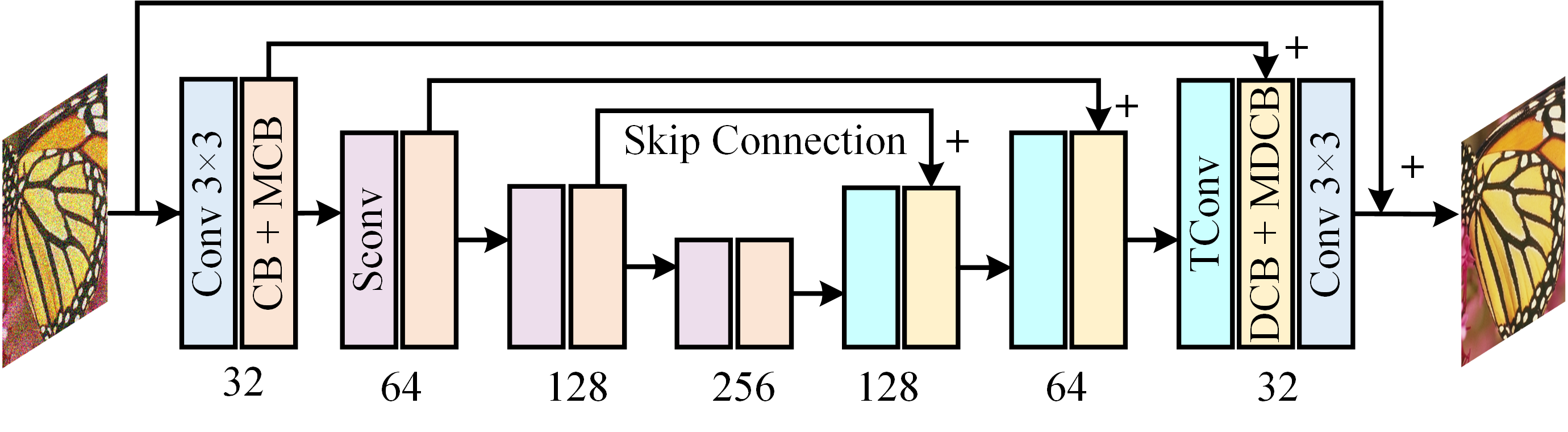} 
		\caption{Illustration of our proposed adaptive dynamic filtering network (ADFNet).}
		\vspace{-0.3cm}
		\label{fig:net}
	\end{figure}

	The architecture of our proposed ADFNet is illustrated in Fig.~\ref{fig:net}. Following~\cite{sadnet}, our ADFNet adopts an encoder-decoder framework to pursue an effective and efficient target. To be detailed, the proposed ADFNet consists of four scales, each except the lowest scale has a residual connection between the encoder and decoder. In the initial stage, we utilize one $3\times3$ convolution (Conv) to extract shallow features from the noisy input. Considering that dynamic convolution is very sensitive to noise, in the encoder, we employ a convolution block (CB) and a multi-scale convolutional block (MCB) to extract features at four scales and filter noise. Then, in the decoder, we reconstruct the features in each scale using a dynamic convolution block (DCB) and a multi-scale dynamic convolution block (MDCB) from coarse to fine. Here, we employ one $3\times3$ strided Conv (SConv) and one $6\times6$ transposed Conv (TConv) for feature downsampling and upsampling, respectively. The number of channels in the encoder and decoder from the first scale to the fourth scale is 32, 64, 128, and 256. 
	
	Without bells and whistles, we optimize the proposed ADFNet using the $L_{2}$ loss for Gaussian image denoising and the Charbonnier loss~\cite{charbonnier} for real-world image denoising. 
	
	\subsection{Spatially Enhanced Kernel Generation (SEKG)}
	As discussed in previous sections, most of previous works discard the spatial information and fail to effectively aggregate channel-specific information. This work attempts to adopt a cost-effective way to achieve kernel generation. 
	
	As shown in Fig.~\ref{fig:drb} (a), our SEKG module is designed with a \textit{spatial context extractor} branch and a \textit{channel information interaction} branch. To reduce the amount of computation and the number of network parameters as much as possible, and to enhance the representation of context information. Inspired by~\cite{bcan} and~\cite{ecanet}, the spatial context extractor branch is implemented with a simple $3\times3$ depthwise separable convolution and the channel information interaction branch contains an average pooling followed by a convolution mapping layer. These two branches effectively exploit the relationship of inter-channel and intra-channel, respectively. In order to make full use of them, we adopt the element-wise addition operation to fuse them. Finally, we utilize one $1\times1$ convolution to output feature maps whose dimension is $(k\times k \times c)\times h \times w$. To formulate spatially-varying kernels, we reshape them into a series of per-pixel kernels $W_{i,j} \in \mathbb{R}^{k^2 \times c}$, where $i \in \{1,2,\cdots,h\}$, $j \in \{1, 2, \cdots, w\}$. In this way, we learn a different high-pass filtering kernel for each location.
	
	\subsection{Dynamic Convolution Block (DCB)}
\begin{figure}[t]
		\centering
		\includegraphics[width=0.45\textwidth]{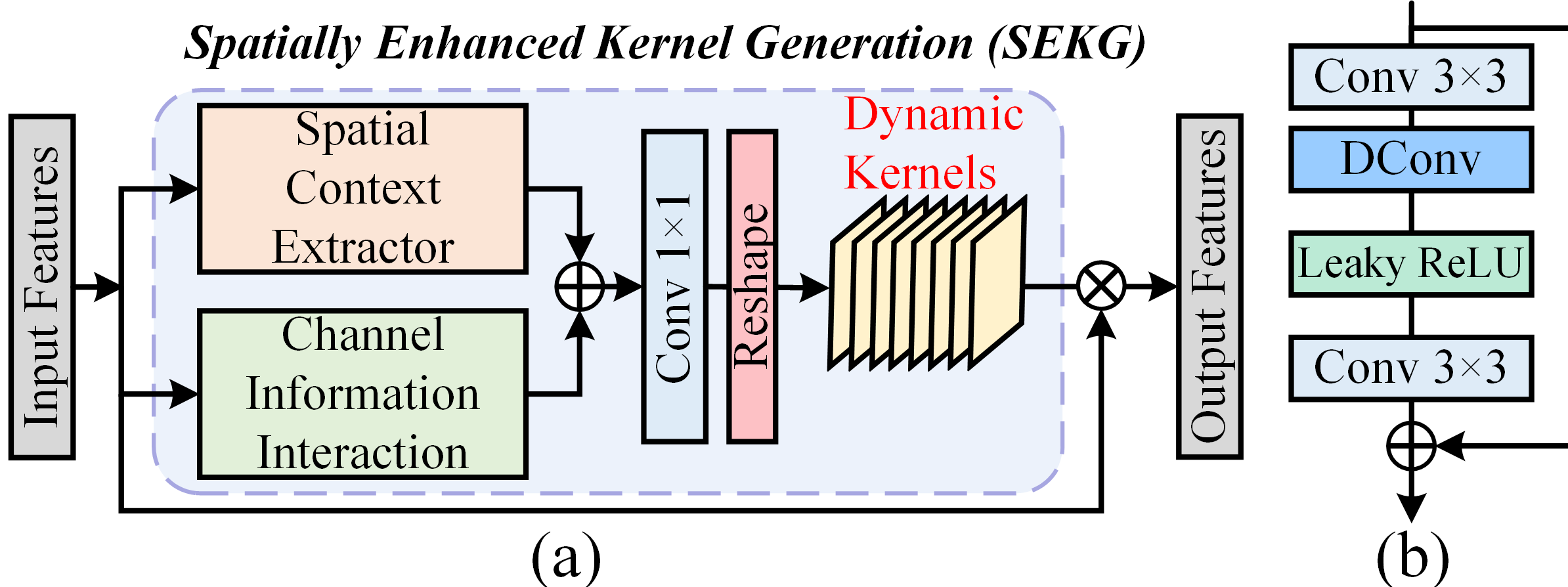} 
		\caption{Illustration of (a): dynamic convolution (DConv) and (b): dynamic convolution block (DCB).}
		\label{fig:drb}
		\vspace{-0.5cm}
	\end{figure}

	It is well-known that high-order operators are more sensitive to disturbance, while the first-order operators are not. In other words, dynamic and plain static convolution can produce complementary representations of various degrees. Therefore, we integrate the proposed dynamic convolution (DConv) into the residual block to take advantage of their own meritc fully. As shown in Fig.~\ref{fig:drb} (b), the DConv is added after the frst convolution in the residual block. Unlike static convolution, which has fixed kernels once trained, dynamic kernels are generated by independent network branches and can be adjusted based on the input. 
	
	\subsubsection{Dynamic Convolution (DConv).}
	We take $W$ as the predicted kernels to perform dynamic convolutional operators on the corresponding input feature maps. It is worth noting that each kernel in $W$ has only $c$ channels, not $c\times c_{out}$, since we utilize the idea of depth-wise separable convolution to decrease the number of parameters further. This manner is completely different from the existing methods~\cite{dfn,cran,hypernetworks} because it actually involves channel-wise adjustment and the operation of each channel is independent. 
	
	\subsection{Multi-Scale Dynamic Convolution Block (MDCB)}
\begin{figure*}[t]
		\centering
		\includegraphics[width=0.8\textwidth]{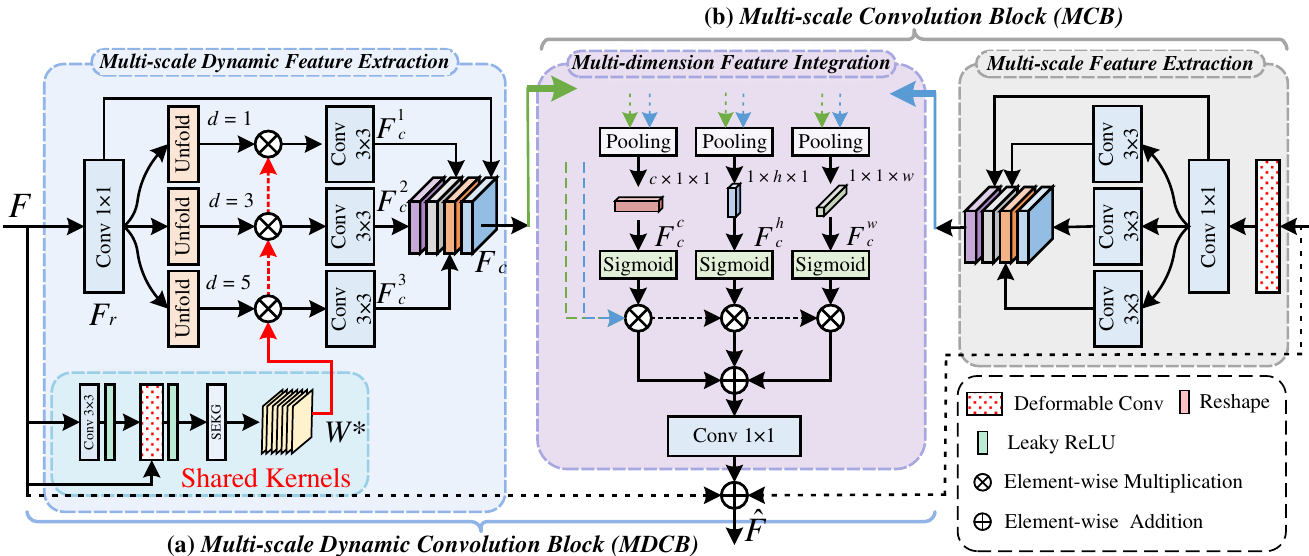} 
		\caption{The detailed design of the proposed multi-scale dynamic convolution block.}
		\label{fig:mdcb}
		\vspace{-0.4cm}
	\end{figure*}

	The above-designed DConv utilizes the content-adaptive properties and spatial context information to guide the dynamic kernel generation locally, thus processing high-frequency information better. Inspired by multi-scale convolution block (MCB) (Fig.~\ref{fig:mdcb} (b)) and dynamic convolution, the MDCB can (1) enlarge the receptive field size; (2) utilize adaptive dynamic kernels to filter multi-scale features globally, thus endowing the model with capabilities to generalize well on diverse noisy input. To this end, the generated kernels should be shared among multi-scale features to reduce computational complexity, which is the main distinction between MCB and our MDCB.
	
	\subsubsection{Multi-Scale Dynamic Feature Extraction.}
	There are two steps to extract multi-scale dynamic features, the first step, named shared adaptive kernel generation, is to generate shared dynamic convolutional kernels. The second aims to aggregate multi-scale features.
	
	\textbf{(1) Shared Adaptive Kernel Generation.} As shown in Fig.~\ref{fig:mdcb} (a), we first use a modulated deformable Conv (MDConv) for powerful feature representation. Then, the SEKG module is embedded for the transformation between features and per-pixel kernels. As we all know, the MDConv can adaptively set the offset of each element and gather values for each component of the local feature patch, thus providing sufficient guidance for features at different positions. In our implementation, the offset is calculated by one $3\times3$ Conv followed by an activation function,
	\begin{equation}
		\{\Delta x, \Delta y, \Delta w\}_{(x, y)}=H\left(F\right),
	\end{equation}
	where $H(\cdot)$ denotes the convolutional operation, $\Delta x, \Delta y$ is the offset with respect to the positions $(x,y)$, and $\Delta w$ is the learnable modulation scalar. Then, the MDConv takes the input features $F$ and learnable offsets as its input to generate the enhanced feature $F_{m}$,
	\begin{equation}
		F_{m}(x, y)={\sum}_{j=1}^{k} w_{j} * F\left(x+\Delta x_{j}, y+\Delta y_{j}\right) * \Delta w_{j},
	\end{equation}
	where $k$ denotes the number of samples, and $w$ is the learnable weights. In this way, we can obtain the fine-grained features and filter unpleasant noises to avoid noise interference with the following kernel generation. Next, we utilize the proposed SEKG module to generate the shared dynamic kernels and denote as $W^{\ast}$. 
	
	\textbf{(2) Multi-Scale Feature Aggregation.} Before transforming features, we selectively to reduce the channel numbers and obtain the reduced features $F_{r}$. Then, we utilize the Unfold operation to extract sliding local 3D patches with patch-size $k=3$, stride $s=1$, and various dilation rates $(d=1,3,5)$ from input features $F_{r}\in\mathbb{R}^{{c}\times{h}\times{w}}$, and then reshape these 3D patches to obtain three groups of new features, denoted as ${F_{t}^{i}}(i=1,2,3)\in\mathbb{R}^{{c}\times{k^2}\times{h}\times{w}}$. As done in the above DConv, we adopt the same process manner to aggregate each scale features. Hence, after concatenating scale-specific features $\{F_{c}^{i}\}_{i=1}^{3}$ and initial features $F_{r}$, the resulted $F_{c}$ can achieve powerful multi-scale feature representations. The entire process can be formulated as
	\begin{equation}
		\begin{aligned}
			F_{c}^{i} &= C_{3\times3}^{i}(F_{t}^{i}\otimes W^{\ast}), \\
			F_{c} &= Concat(F_{r}, F_{c}^{1}, F_{c}^{3}, F_{c}^{3}),
		\end{aligned}
	\end{equation}
	where $\otimes$ denotes the element-wise multiplication operation, $C_{3\times3}^{i}(i=1,2,3)$ is three independent $3\times3$ Convs which are used to interwove with the dynamic multi-scale features, $Concat(\cdot)$ is the channel concatenation function, and $F_{c}$ denotes the final output. 
	
	\begin{table*}[htbp]
		\centering
		\resizebox{0.99\textwidth}{!}{
			\begin{tabular}{|c|c|c|c|c|c|c|c|c|c|c|c|} 
				\hline
				\multicolumn{1}{|c|}{Datasets}      
				& \makecell[c]{Noise \\ Level}
				& \makecell[c]{DnCNN}
				& \makecell[c]{FFDNet}	
				& \makecell[c]{RNAN}     
				& \makecell[c]{RIDNet}
				& \makecell[c]{RDN}
				& \makecell[c]{SADNet}   	
				& \makecell[c]{DeamNet}     
				& \makecell[c]{P3AN} 
				& \makecell[c]{MSANet} 
				& \makecell{\textbf{ADFNet}}     \\ \hline \hline
				\multicolumn{1}{|c|}{\multirow{3}{*}{Kodak24}}  & 30 	 	& 31.39		& 31.39     & 31.86    & 31.64    & 31.94    & 31.86    & 31.88   & 31.99    & 31.78	& \textbf{32.01}     \\ 
				\multicolumn{1}{|c|}{}                          & 50 	 	& 29.16		& 29.10     & 29.58    & 29.25    & 29.66    & 29.64    & 29.70   & 29.69    & 29.57    & \textbf{29.81}     \\
				\multicolumn{1}{|c|}{}                          & 70 	 	& 27.64		& 27.68     & 28.16    & 27.94    & 28.20    & 28.28    & 28.30   & 28.25    & 28.17    & \textbf{28.48}     \\ \hline	
				\multicolumn{1}{|c|}{\multirow{3}{*}{BSD68}}    & 30 	 	& 30.40		& 30.31     & 30.63	   & 30.47	  & 30.67    & 30.64    & 30.70	  & 30.72    & 30.67    & \textbf{30.74}     \\ 
				\multicolumn{1}{|c|}{}                          & 50 	 	& 28.01		& 27.96     & 28.27    & 28.12    & 28.31    & 28.32    & 28.34	  & 28.37    & 28.36    & \textbf{28.43}    \\
				\multicolumn{1}{|c|}{}                          & 70 	 	& 26.56		& 26.53     & 26.83    & 26.69    & 26.85    & 26.93    & 26.95   & 26.94    & 26.96    & \textbf{27.03}     \\ \hline 
				\multicolumn{1}{|c|}{\multirow{1}{*}{MCMaster}} & 50 	 	& 28.62	    & 29.18     & 29.18    & -        & 29.60    & 29.72	& 29.78   & -        & 29.82    & \textbf{29.96} \\ 
				\hline
		\end{tabular}}
		\caption{Average PSNR (dB) results of different methods for color image denoising on various datasets.}
		\label{tab:color_table}
	\end{table*}
	\begin{figure*}[!htbp]
		\centering
			\subfigure[Noisy]{\includegraphics[width=0.16\textwidth]{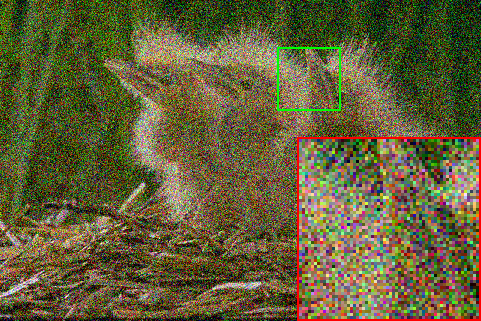}}
			\subfigure[DnCNN]{\includegraphics[width=0.16\textwidth]{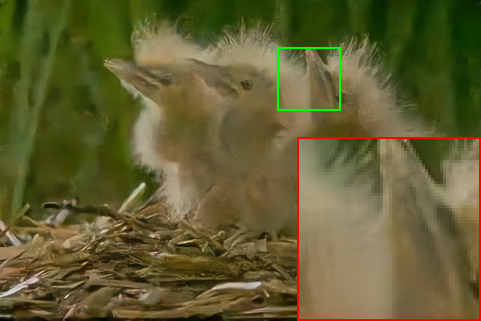}}
			\subfigure[RNAN] {\includegraphics[width=0.16\textwidth]{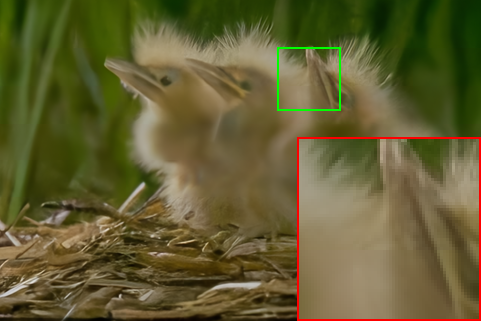}}
			\subfigure[SADNet]{\includegraphics[width=0.16\textwidth]{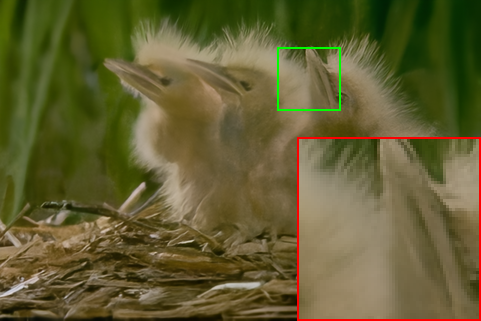}}
			\subfigure[DeamNet]{\includegraphics[width=0.16\textwidth]{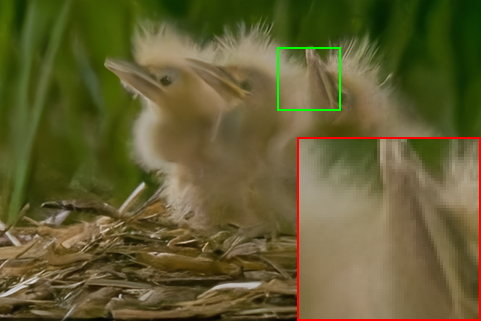}}
			\subfigure[ADFNet]{\includegraphics[width=0.16\textwidth]{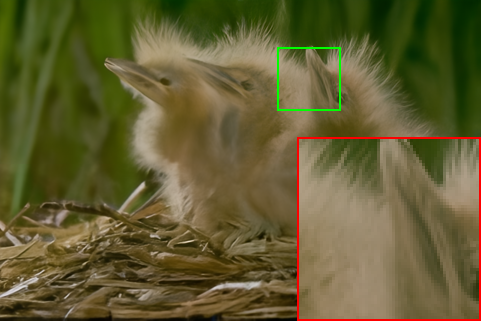}}
		\vspace{-0.3cm}
		\caption{Visual comparisons of various methods for color gaussian image denoising. The noisy image is corrupted by additive white gaussian noise (AWGN) with noise level $\sigma=50$.}
		\label{fig:vis} 
	\end{figure*}
	
	\subsubsection{Multi-Dimension Feature Integration.}
	Regarding feature integration, inspired by the visual attention mechanism, we perform integration through a multi-dimension attention-guided mechanism. Specifically, we obtain three groups of contextual information by average-pooling operation on different dimensions of feature maps. We use channel-wise attention ($(h,w)$ dimension) as an instance. An average-pooling operation along the channel axis is firstly employed on $F_{c}$ to generate an efficient feature descriptor. Based on the feature descriptor, we then apply a sigmoid function to generate a spatial attention map ranging from 0 to 1, which is used to encode where to highlight or suppress. Finally, we perform an multiplication operation between $F_{c}$ and descriptor to obtain spatial-wise attentive features $F_{c}^{c}$. Likewise, we can obtain another two attention maps from $(c, h)$ and $(c, w)$ dimensions. To fully take advantage of them and explore the cross-dimension interaction between channel and spatial contexts, we take the above obtained $F_{c}^{c}$ as the calibrated features and then multiply them with other two attention maps, respectively, to acquire two new attentive features $F_{c}^{w}$ and $F_{c}^{h}$. Formally, we have
	\begin{equation}
		\begin{aligned}
			F_{c}^{c} &= Sigmoid(AvgPool(F_{c}))\otimes F_{c}, \\
			F_{c}^{w} &= Sigmoid(AvgPool(F_{c}))\otimes F_{c}^{c}, \\
			F_{c}^{h} &= Sigmoid(AvgPool(F_{c}))\otimes F_{c}^{c}. \\
		\end{aligned}
	\end{equation}
	Finally, we add three attentive features and employ a convolutional layer to fuse them.
	\begin{equation}
		\hat{F} = F + C_{1\times1}(F_{c}^{c} + F_{c}^{w} + F_{c}^{h}),
	\end{equation}
	where $C_{1\times1}$ denotes the $1\times1$ Conv, $\hat{F}$ is the final output of the MDCB. Here, we adopt local residual learning to further improve the information flow.
	
	By capturing the cross-dimension inter-dependencies, the extracted multi-scale dynamic features can be fused further and show a more powerful representative capability.
	
	\section{Experiments}
	\subsection{Experimental Setup}
	\textbf{Datasets.} In this work, we use 800 images from DIV2K~\cite{div2k} as training data for Gaussian image denoising. For real image denoising, the Smart-phone Image Denoising Data set (SIDD)~\cite{sidd} is chosen to train models. There are 30K real noisy images in the SIDD, 320 image pairs for training, and 1280 images for validation. We randomly crop $320\times300$ image patches with a size of $256\times256$ to train the model due to the large size of these primitive real images.
	
	For Gaussian image denoising, we employ BSD68~\cite{bsd68}, Kodak24~\cite{kodak24}, and McMaster~\cite{McMaster} as test sets. For real image denoising, Darmstadt Noise Dataset (DND)~\cite{dnd} and SIDD~\cite{sidd} (\eg benchmark and validation sets) are chosen for testing. It is worth noting that due to the ground truths of the real images are not publicly available at present, we obtain the results of quantitative assessment via the online submission system.
	\newline
	\textbf{Implementation Details.}
	We implement ADFNet in PyTorch framework and conduct all experiments on NVIDIA 2080-Ti GPUs. During training, we employ ADAM~\cite{adam} optimizer that the parameters are set as $\beta_{1}$ = 0.9, $\beta_{2}$ = 0.999, $\epsilon=10^{-8}$ to learn the optimal parameters. We randomly crop 16 patches of size $128\times128$ as input for each training mini-batch and augment the patches by flipping horizontally or vertically and rotating \ang{90}. The learning rate is initialized to ${10}^{-4}$ and then linearly decreases to half every $2\times10^{5}$ iterations. 
	
	\begin{table}[t]
		\centering
		\resizebox{0.45\textwidth}{!}{
			\begin{tabular}{|l|c|c|c|c|}
				\hline
				Methods                 & FLOPs    & Time    & Kodak24       & McMaster\\ \hline \hline
				SwinIR*    			    & 186      & 0.472       & 29.79         & 30.22 \\ \hline
				MalleNet*   			& -        & -           & 29.61         & \textbf{30.23} \\ \hline
				\textbf{ADFNet}         & \textbf{13.9}     & \textbf{0.027}       & 29.81         & 29.96 \\ \hline
				\textbf{ADFNet*}        & 46.2     & 0.057       & \textbf{29.92}         & 30.12 \\ \hline
		\end{tabular}}
		\caption{Comparisons of performance and model complexity between SOTA methods. `*' denotes models trained on DIV2K + Flickr2k + WED + BSD400 datasets.}
		\vspace{-0.3cm}
		\label{tab:color_supp}
	\end{table}
	\begin{table*}[ht]
		\centering
		\resizebox{0.98\textwidth}{!}{
			\begin{tabular}{|c|c|c|c|c|c|c|c|c|c|c|c|}
				\hline
				Datasets 
				& \makecell[c]{DnCNN}  
				& \makecell[c]{CBDNet*} 
				& \makecell[c]{RIDNet} 
				& \makecell[c]{AINDNet*} 
				& \makecell[c]{VDN}    
				& \makecell[c]{MPRNet} 
				& \makecell[c]{DeamNet*} 
				& \makecell[c]{DAGL}  
				& \makecell[c]{MSANet*}
				& \makecell[c]{\textbf{ADFNet}} \\ \hline \hline
				\multirow{2}[0]{*}{\makecell[c]{SIDD \\ Validation}} 
				& 38.56  	& 38.68  	& 38.71  	& 39.08  	& 39.28   	& 39.71  			& 39.47  	& 38.94  	& 39.56		& \textbf{39.79}  \\
				& 0.910  	& 0.909  	& 0.951  	& 0.954  	& 0.956   	& 0.958     		& 0.955  	& 0.953  	& 0.912			& \textbf{0.960}  \\ \hline
				\multirow{2}[0]{*}{\makecell[c]{SIDD \\ Benchmark}} 
				& 23.66  	& 33.28  	& -      	& 38.08  	& 39.26   	& -      		    & 39.35  	& -      	& -         & \textbf{39.63}  \\
				& 0.583  	& 0.868  	& -      	& 0.953  	& 0.955   	& -      			& 0.955  	& -      	& -         & \textbf{0.958}  \\ \hline
				\multirow{2}[0]{*}{DND} 
				& 32.43  	& 38.06  	& 39.26  	& 39.53  	& 39.38   	& 39.80  			& 39.63  	& 39.77  	& 39.65     & \textbf{39.87}  \\
				& 0.790  	& 0.942  	& 0.953  	& 0.956  	& 0.952   	& 0.954  			& 0.953  	& \textbf{0.956} & 0.955    	& 0.955  \\
				\hline
		\end{tabular}}
		\caption{Average PSNR (dB) and SSIM results of different methods for real-world image denoising on DND dataset and SIDD benchmark datasets. `*' denotes models trained on additional traing sets.}
		\label{tab:real}
	\end{table*}
	\subsection{Color Gaussian Image Denoising}
	To verify the effectiveness of the proposed ADFNet for Gaussian noisy images, we compare it with the existing methods include DnCNN~\cite{dncnn}, FFDNet~\cite{ffdnet}, RNAN~\cite{rnan}, RIDNet~\cite{brdnet}, RDN~\cite{rdn}, SADNet~\cite{sadnet}, DeamNet~\cite{deamnet}, P3AN~\cite{p3an}, and MSANet~\cite{msanet}.
	The compared PSNR values are listed in Table~\ref{tab:color_table}. From the quantitative results, we observe that our ADFNet consistently achieves better PSNR values on all datasets. Taking the Kodak24 dataset as an example, our model obtains a performance gain of 0.13 dB, 0.11 dB, and 0.18 dB over the DeamNet on three different noise levels. More importantly, Deamnet employs an additional BSD400 dataset to train the model. These persistent performance improvements also demonstrate that our ADFNet has powerful capabilities to restore strong corrupted images. Besides, some CNN- and Transformer-based approaches~\cite{swinir,mallConv} enrich training sets to boost model performance. As such, we also utilize more data to train a novel model named \textbf{ADFNet*}, which is built by setting the channel numbers from the first to the last scale as 64, 128, 256, and 512. From Table~\ref{tab:color_supp}, we can see the SwinIR~\cite{swinir} takes more than $\times$\textbf{8} time and $\times$\textbf{4} FLOPs to process one $128\times128$ image compared to our model. However, the PSNR value on Kodak24 is still lower than ours. Although our methods cannot achieve the best results on all benchmarks, the superiority of inference time and computational complexity are apparent.
	
	
	We visually compare our denoised images with some competitive approaches. As shown in Fig.~\ref{fig:vis}, the image generated by our ADFNet preserves rich details, while RNAN and DeamNet introduce some smoothness. 
	
	\subsection{Real-World Image Denoising}
\begin{figure}[t]
		\centering
		\subfigure[]	
		{
			\includegraphics[width=0.472\linewidth]{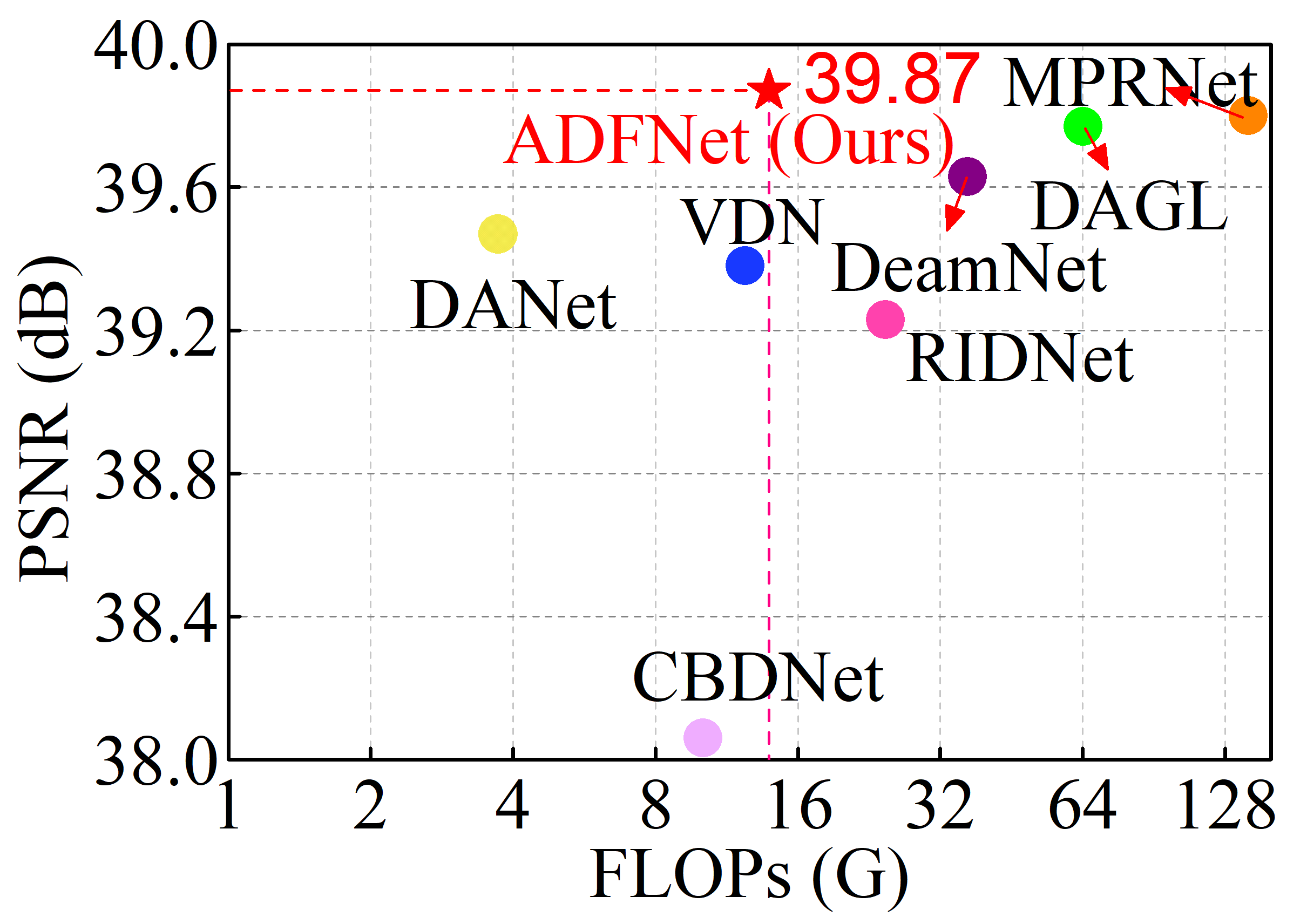}
			\label{fig:time}
		}
		\subfigure[]	
		{
			\includegraphics[width=0.472\linewidth]{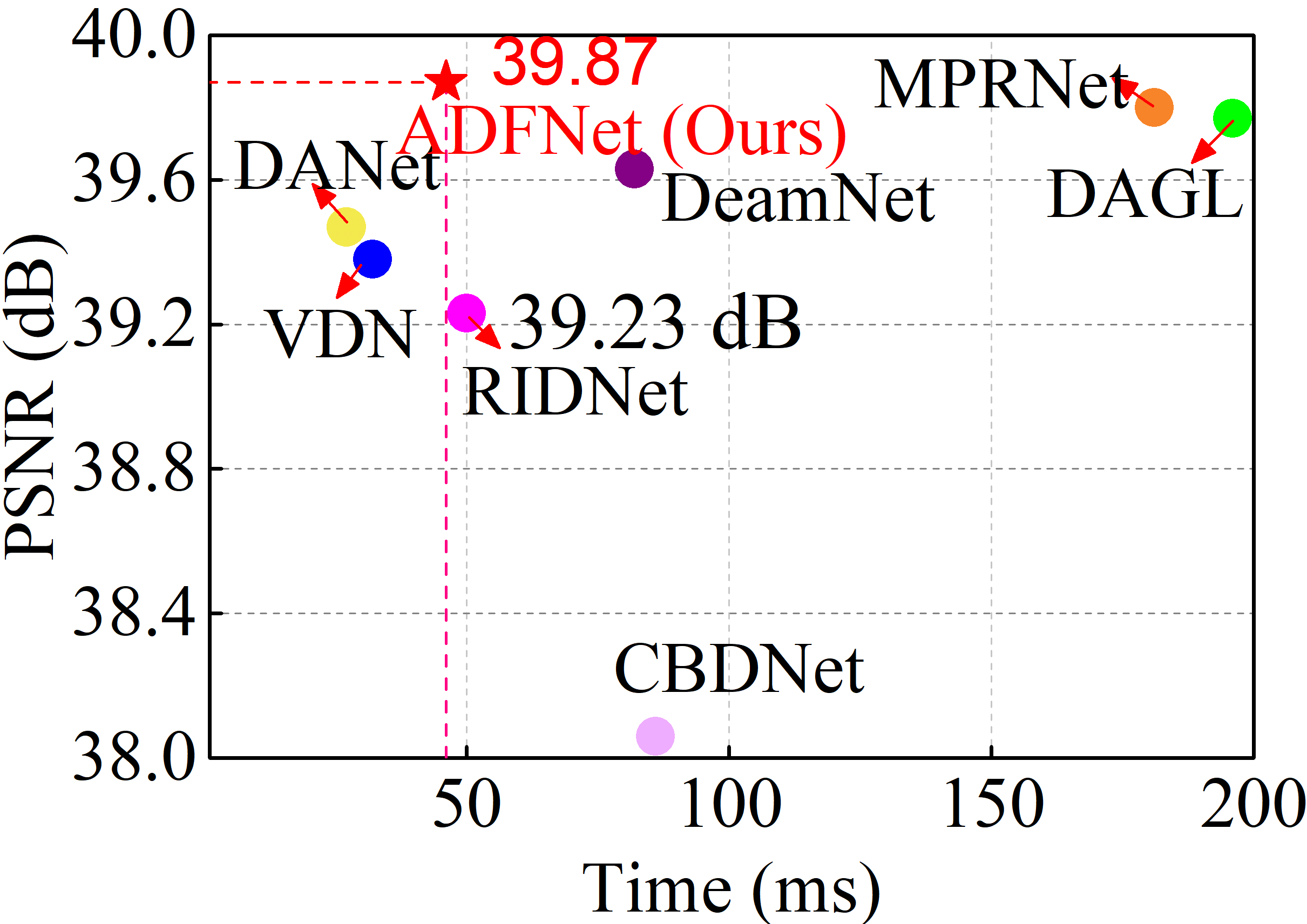}
			\label{fig:flop}
		}
		\vspace{-0.6cm}
		\caption{Trade-off between the performance, FLOPs and running time for real-world image denoising. The PSNR scores are evaluated on the DND dataset and the FLOPs are obtained by calculating one 128$\times$128 patch.}
		\vspace{-0.6cm}
		\label{fig:time_flop}
	\end{figure}

	We compare our ADFNet with recent leading methods for real-world image denoising, such as DnCNN~\cite{dncnn}, CBDNet~\cite{cbdnet}, RIDNet~\cite{ridnet}, AINDNet~\cite{aindnet}, VDN~\cite{vdn}, MPRNet~\cite{mprnet}, DeamNet~\cite{deamnet}, DAGL~\cite{dagl}, and MSANet~\cite{msanet}. The results of different methods are provided in Table~\ref{tab:real}. We can conclude that ADFNet achieves the best PSNR results on three benchmarks, including the newly proposed DAGL and MSANet. It is worth noting that some methods train models with more training sets but achieve lower performance. Overall, the performance superiority of ADFNet confirms the validity and rationality of our network design. \newline
	\textbf{Evaluation on Model Complexity.}
	We evaluate the model complexity of some representative methods in the last three years for real-world image denoising. The FLOPs and running time are chosen as the primary evaluation metrics. The FLOPs are calculated on one 128$\times$128 resolution RGB patch. The running time is evaluated on the DND dataset, and each result is obtained by repeating five experiments to ensure a fair comparison. It can be seen from Fig.~\ref{fig:time} that DANet~\cite{danet} has the lowest FLOPs, but the performance is far worse than our method. MPRNet~\cite{mprnet} is close to the performance of our ADFNet, but the FLOPs cost is more than $\times$\textbf{10} ours. Next, we study the trade-off between the running time and the performance of various methods. As shown in Fig.~\ref{fig:flop}, RIDNet~\cite{ridnet} has a similar execution time compared to our method, but our ADFNet surpasses its performance by a large margin from \textbf{39.23 dB} to \textbf{39.87 dB}. Further, compared to the top methods like MPRNet and DAGL~\cite{dagl}, we have a slight performance improvement in the PSNR but run nearly $\times$\textbf{4} faster. Therefore, compared with these approaches, our ADFNet achieves a better trade-off between performance and FLOPs cost.
	\begin{figure*}[htbp]
		\centering
		\includegraphics[width=0.48\linewidth]{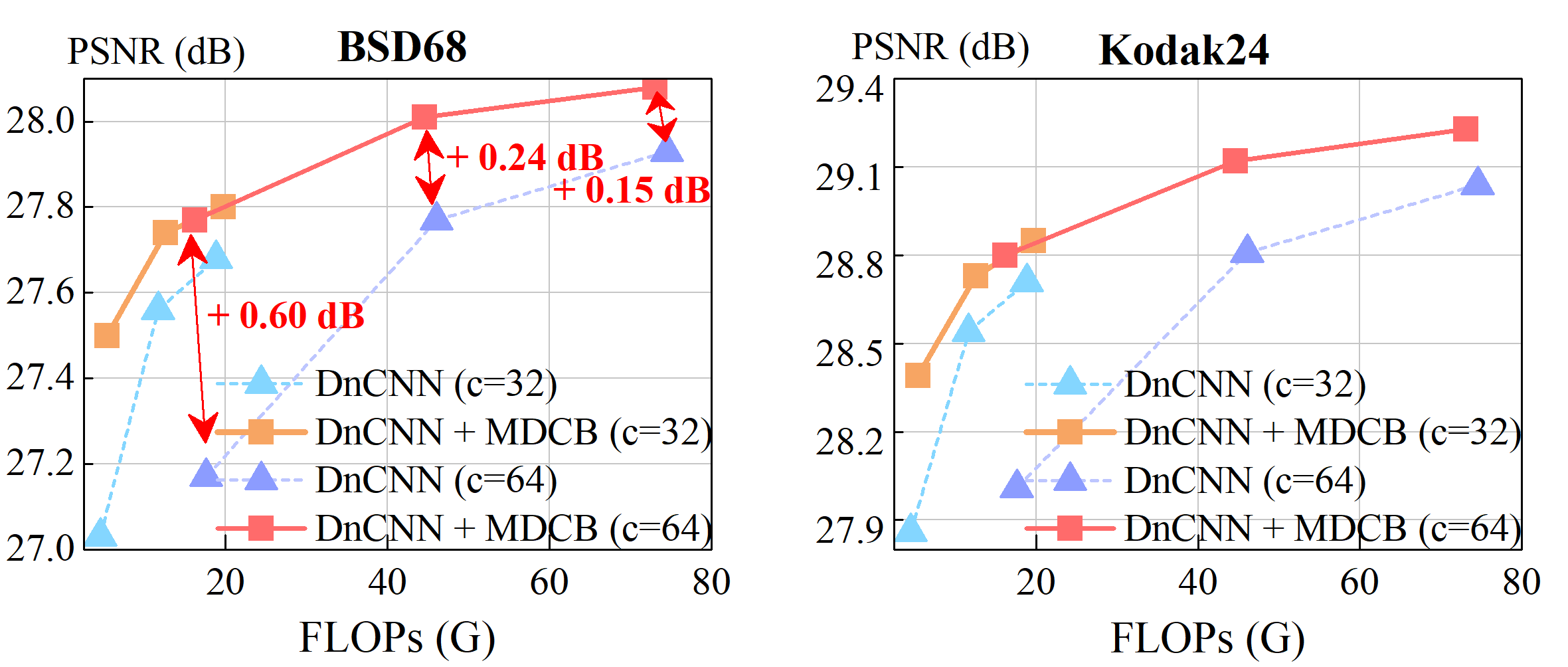}
		\includegraphics[width=0.48\linewidth]{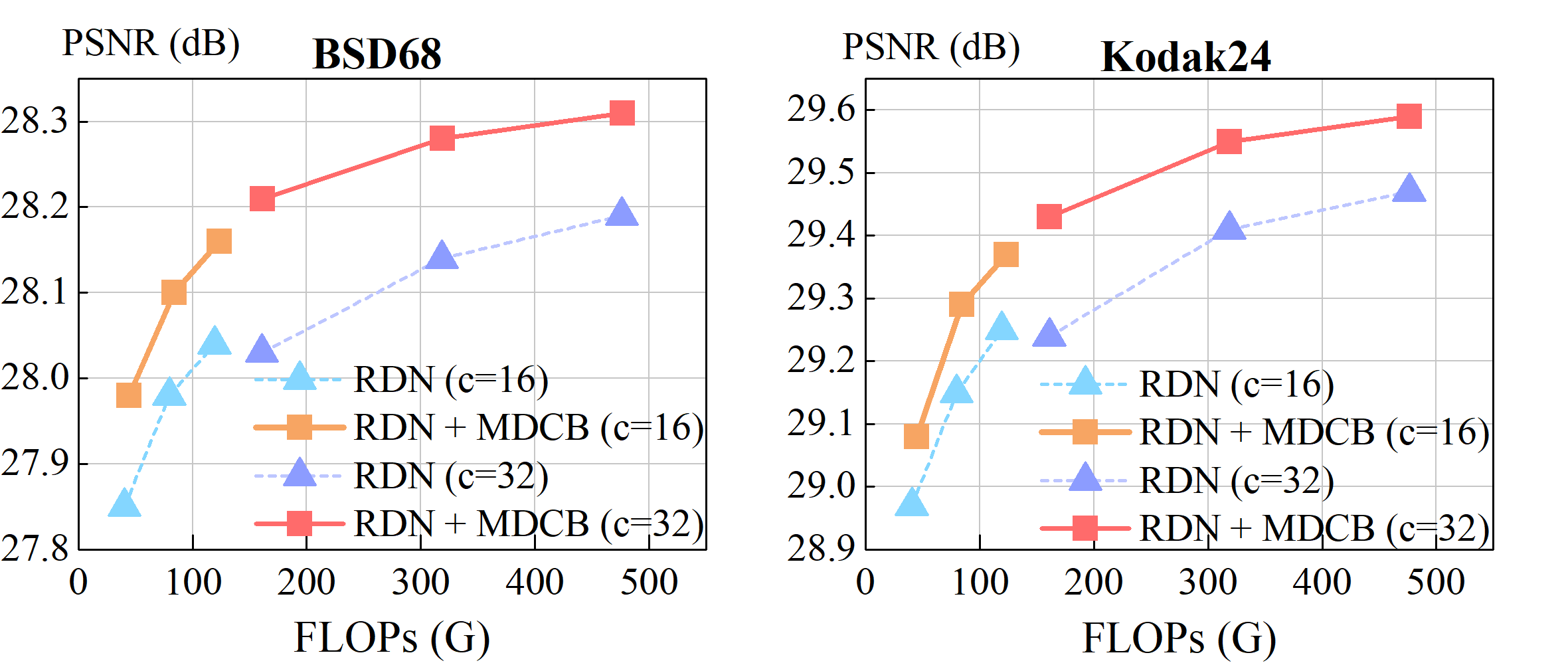}
		\caption{Applying DConv and MDCB to different backbones. The DConv-enhanced DnCNN is built by setting depth = (5, 10, 15) and channel = (32, 64). The MDCB-enhanced RDN is built by setting block number = (3, 6, 9) and channel = (16, 32).}
		\vspace{-0.5cm}
		\label{fig:backbone_res}
	\end{figure*}
	\subsection{Extension: DConv and MDCB Used in Backbones}
	To further demonstrate the effectiveness of the proposed DConv and MDCB, we conduct a group of experiments by plugging them into existing well-designed network architectures. Here, we mainly choose two representative methods, DnCNN~\cite{dncnn} and RDN~\cite{rdn}, as the backbone networks for color Gaussian image denoising. Specifically, we manually build novel networks based on the original two network architectures by adjusting the number of channels and network depth. For instance, we design lightweight and effective DnCNN by setting the depth = (5, 10, 15) and channel = (32, 64). Likewise, we implement deep network RDN by controlling block numbers as (3, 6, 9) and channels as (16, 32). As comparisons, we reimplement the DConv-enhanced DnCNN by replacing one regular static $3\times3$ convolution with one DConv in the middle of the network. The MDCB-enhanced RDN is developed by replacing one residual dense block with three MDCBs in the middle of the backbone network. 
	
	As shown in Fig.~\ref{fig:backbone_res}, the DConv-enhanced DnCNN achieves performance gains by 0.60 dB, 0.24 dB, and 0.15 dB, respectively, with the increases in the network depth. According to these observations, we can draw the following conclusions. First, the DConv- and MDCB- enhanced corresponding versions can achieve better performance in the lightweight and deep networks. Second, the performance superiority is decreased with the increases of the channels and depth, but stacking more modules may bring adequate performance improvements further. Third, the SEKG-guided modules enhance the representational capabilities of networks and can serve as a normal layer or an essential feature extraction module, mainly attributed to the generated kernels fully utilizing the spatial context information.
	
	\begin{table}[htbp]
		\centering
		\resizebox{0.47\textwidth}{!}{
			\begin{tabular}{|c|c|c|c|}
				\hline
				Model                     	   & Params	   & Kodak24 & BSD68 	 \\ \hline \hline
				w/o (DCB and MDCB)        & 7.71	   & 29.61   & 28.33 	    \\ \hline
				w/o DCB 	      	    & 7.64      & 29.69   & 28.36 	    \\ \hline
				w/o MDCB         	    & 7.72      & 29.66   & 29.37 	    \\ \hline
				w/o MFI 	      	    & 7.65      & 29.77   & 28.40 	    \\ \hline
				\makecell{w/ (DCB \& MDCB) (all)}      & \textbf{7.30}      & 29.77   & 28.38 	   \\  \hline
				\makecell{w/ (DCB \& MDCB) (encoder)}     & 7.65      & 29.73   & 28.38 	    \\  \hline
				\makecell {w/ (DCB \& MDCB) \\ (decoder) ADFNet}           	    & 7.65      & \textbf{29.81}   & \textbf{28.43}     \\ \hline
		\end{tabular}}
		\caption{Ablation study on different components.}
		\label{tab:ablation}
		\vspace{-0.5cm}
	\end{table}
	
	\subsection{Ablation Studies}
	We study the impact of each component on the performance of our ADFNet. All experiments are performed on the color Gaussian image denoising with noise level $\sigma$ = 50. The ablation models are trained on image patches of the size of $128\times128$ with the same number of iterations.  
	
\begin{figure}[t]
		\centering
		\includegraphics[width=0.9\linewidth]{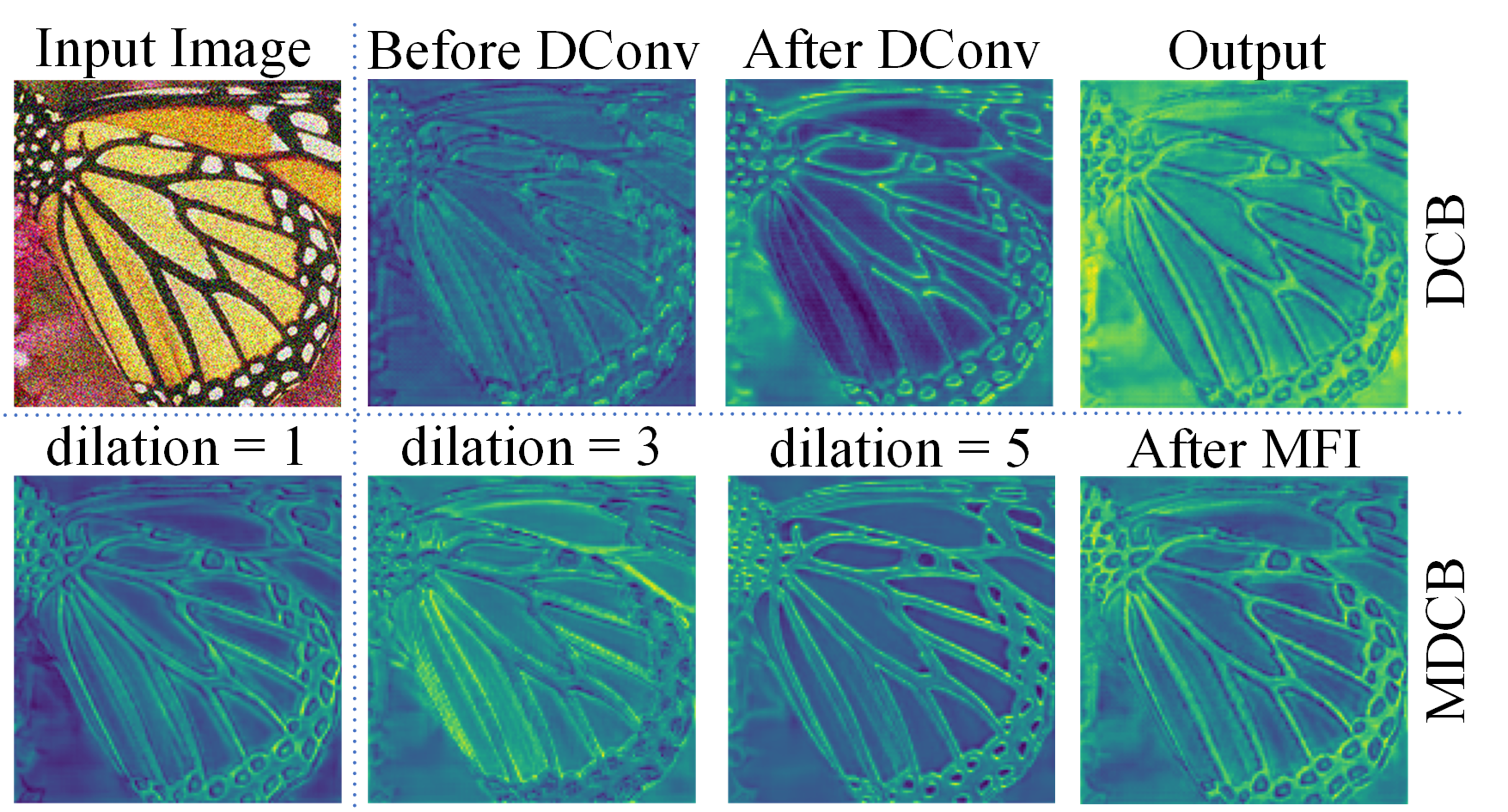}
		\caption{Visualization of feature maps extracted by the first DCB and MDCB and their inner convolutions.}
		\vspace{-0.6cm}
		\label{fig:freq}
	\end{figure}

	\textbf{Effectiveness of SEKG Module.}
	As previously mentioned, the proposed DCB and MDCB are designed based on the SEKG module. To demonstrate the validity of the SEKG, we replaced the DCB with a regular $3\times3$ convolution block, denoted as \textit{\textbf{w/o DCB}}. From Table~\ref{tab:ablation}, we see the model has a significant performance degradation on all benchmarks, but the number of parameters and FLOPs is almost the same as our ADFNet. We also replace the MDCB in the decoder with a static multi-scale convolution block (Fig.~\ref{fig:mdcb}). This block is composed of a modulated deformable convolution and three parallel $3\times3$ convolutions with different dilation rates, and the main distinction lies in the kernels of our MDCB generated by an individual kernel generation module. As shown in Table~\ref{tab:ablation}, we observe the \textbf{\textit{w/o MDCB}} obtains worse performance, while our ADFNet obtains better results. The performance degrades further when the DCB and MDCB are replaced simultaneously, indicating that the proposed SEKG promotes kernel generation. To have better insight, we visualize the average feature maps after applying the DCB and MDCB. From the Fig.~\ref{fig:freq}, we have two observations. First, the DConv used in DCB is able to highlight the high-frequency information, and the low-frequency information can be incorporated via skip connections. Second, the shared dynamic kernels used in MDCB can enhance the multi-scale feature representations. These indicate that the joint low- and high-frequency information structure is reasonable to deploy in the decoder to reconstruct features.

	Since dynamic convolution is very sensitive to noise, we only deploy the proposed modules in the decoder. When the DCB and MDCB are deployed in the phase of encoding, the performance is lower than our proposed scheme. This phenomenon is consistent with our analysis in the previous section, that is, noisy kernels are generated during the encoder period, and more robust kernels are generated in the decoding phase with the removal of noise. 
	
	\textbf{Effectiveness of MFI Mechanism.} To evaluate the effectiveness of the module MFI, we remove it from the architecture of our ADFNet for comparison, denoted as \textbf{\textit{w/o MFI}}. Thus, the produced dynamic multi-scale features are concatenated along the channel dimension but omit the fully multi-dimension feature fusion. From Table~\ref{tab:ablation}, we see an apparent degradation in terms of PSNR score on three datasets, indicating the MFI can accurately discover the correlation of multi-scale features and complete the information interaction among different feature dimensions to achieve better representations. Fig.~\ref{fig:freq} shows the feature maps after employing the MFI mechanism in MDCB, and we can see the attention mechanism has a noticeable effect on modulating the activation values. These results demonstrate the effectiveness of the fusion mechanism.
	
	\section{Conclusions}
	In this work, we propose a spatially enhanced kernel generation (SEKG) module to improve the dynamic convolution while maintaining low computational costs. With the proposed module, we further present a dynamic convolution block and a multi-scale dynamic convolution block to improve the representations of high-frequency and multi-scale features. Therefore, benefiting these modules, the proposed ADFNet can effectively remove noise and preserve fine-grained image details as verified in experiments. 

	\section{Acknowledgments}
	This work was supported in part by the National Natural Science Foundation of China under Grant 61976079, in part by Anhui Key Research and Development Program under Grant 202004a05020039, and in part by the Key Project of Science and Technology of Guangxi under Grant AB22035022- 2021AB20147.

	\bibliography{aaai23}
\end{document}